\documentclass{ecai}
\usepackage{times}
\usepackage{graphicx}
\usepackage{latexsym}
\usepackage{algorithm,algorithmic}
\usepackage{booktabs}

\ecaisubmission   

\begin{document}

\title{Adversarial Robustness for Machine Learning Cyber Defenses Using Log Data}

\author{Kai Steverson\institute{DCI-Solutions; email: ksteverson@dci-solutions.com} \and Jonathan Mullin\textsuperscript{1} \and Metin Ahiskali\institute{U.S. Army Combat Capabilities Development Command (CCDC),		
C5ISR Center} }

\maketitle
\bibliographystyle{ecai}

\begin{abstract}
There has been considerable and growing interest in applying machine learning for cyber defenses. One promising approach has been to apply natural language processing techniques to analyze logs data for suspicious behavior. A natural question arises to how robust these systems are to adversarial attacks. Defense against sophisticated attack is of particular concern for cyber defenses. In this paper, we develop a testing framework to evaluate adversarial robustness of machine learning cyber defenses, particularly those focused on log data. Our framework uses techniques from deep reinforcement learning and adversarial natural language processing. We validate our framework using a publicly available dataset and demonstrate that our adversarial attack does succeed against the  target systems, revealing a potential vulnerability. We apply our framework to analyze the influence of different levels of dropout regularization and find that higher dropout levels increases robustness. Moreover 90\% dropout probability exhibited the highest level of robustness by a significant margin, which suggests unusually high dropout may be necessary to properly protect against adversarial attacks. 
\end{abstract}

\section{INTRODUCTION}

There has been considerable and growing interest in applying machine learning for cyber defenses \cite{buczakSurveyDataMining22,chalapathyDeepLearningAnomaly2019}. This interest is not only academic but has found its way to commercial defensive tools as well. One promising approach is to apply natural language processing techniques to various kind of log data, such as operating system or server logs  \cite{brownRecurrentNeuralNetwork2018,duDeepLogAnomalyDetection2017,tuorRecurrentNeuralNetwork,tuorDeepLearningUnsupervised,yinDeepLearningApproach2017}. These algorithms use an unsupervised anomaly detection approach that first learn a baseline of normal behavior for a set of logs and then flag as suspicious any behavior that deviates from this baseline. 
 
As these approaches become more common, a natural question arises to how robust these systems are to adversarial attacks, particularly offensive uses of machine learning. Adversarial robustness has been a growing concern across machine learning \cite{goodfellowExplainingHarnessingAdversarial2015,szegedyIntriguingPropertiesNeural2014}, but these concerns are particularly sharp in cyber defense where sophisticated attacks are expected. And there has indeed been a growing body of research into adversarial robustness of cyber defenses that use machine learning  \cite{huBlackBoxAttacksRNN,linIDSGANGenerativeAdversarial2019,usamaGenerativeAdversarialNetworks2019}, although none that focus on log data. These adversarial attacks train a generative model to produce data that fool the cyber detector, which allows malicious behavior to pass undetected.

In this paper, we develop a adversarial attack algorithm to evaluate the robustness of machine learning anomaly detection models on log data. Our attack uses techniques from deep reinforcement learning \cite{suttonPolicyGradientMethods2000,williamsSimpleStatisticalGradientfollowing1992} and  adversarial natural language processing \cite{liAdversarialLearningNeural2017,yuSeqGANSequenceGenerative2017}. Using publicly available data, we demonstrate that this type of adversarial attack does work, revealing a potential vulnerability. We then apply our framework to analyze the influence of different levels of dropout regularization. We find that robustness against adversarial attack increases with higher dropout levels. Moreover, the most robustness by a significant margin occurs at 90\% drop probability, which is well outside the usual range for the models we use. This suggests counter-intuitively high levels of dropout may be required to protect against adversarial attacks. 

Our adversarial attack focuses on the case where the attacker has acquired a black-box copy of the detector model and uses its feedback to train a generative model that can fool the detector. By ``black-box", we mean the attacker does not have access to the internal workings of the detector, such as model parameters and learned weights, since that information may be encrypted or otherwise inaccessible. Instead, the attacker can only use the detector to get anomaly scores for different inputs. The attacker could acquire the detector by either stealing it or using live feedback from the detector in its operational environment. The second case assumes the attacker can observe the live output of the detector and had some ability to manipulate inputs on the system. 

Another key feature of our approach is we do not assume the attacker has access to the data on which the detector was trained, or any data closely matching it. We believe this increases realism in the case of attacks against unsupervised anomaly detection models. Those unsupervised models are trained to understand specific patterns of the networks where they are deployed. Hence, getting representative data means getting data from those specific devices, which can be quite large and decentralized. 

Our attack framework does allow the attacker access to a small pretraining dataset that only loosely resembles the type of data the detector expects to ingest. This pretrain dataset is constructed based on the attacker's best guess of what the target system looks like. For example, a loosely representative set of operating systems logs can be generated by general familiarity with how that operating systems works.  Alternatively, data that is out of date or from an emulated environment could be used. We assume this pretrain data is not sufficient to train a useful model, but is instead used to learn some basic structure and a partial vocabulary. In our experiments, we construct the pretrain data using a small subset of the data and adding significant noise by shuffling the data within each column. Our experimental results demonstrate that this pretrain data is not enough to train a useful model on its own. 
 
We test our adversarial attack algorithm using publicly available data from Los Alamos National Lab (LANL) \cite{kentComprehensiveMultiSourceCyberSecurity2015,kentCybersecurityDataSources2015}.\footnote{Data available at https://csr.lanl.gov/data/cyber1/.} The data contains authentication logs from their internal networks collected over the span of 58 days. We train unsupervised anomaly detection models by following previous work by Brown et al. \cite{brownRecurrentNeuralNetwork2018} on the same dataset. We train six such anomaly detectors with different levels of dropout regularization (0\%, 10\%, 30\%, 50\%, 70\%, and 90\%). We then apply our adversarial attack algorithm to each of these six models to determine which is the most robust to adversarial attack. 

Our results demonstrate that our adversarial attack produces a generative model that can successfully fool the detector, revealing a potential vulnerability. We also show that the robustness against this type of attack increases with higher levels of dropout regularization in the detector. Robustness was especially increased at dropout of 90\%, which is well above the level typically used. This implies that unusually high dropout levels increases robustness may be necessary to defend against an adversarial attack. Additionally, higher dropout models perform worse on the baseline detection task without an adversary. Hence, there appears to be a trade-off between the most robust model and those that perform best at their baseline task, which is in line with what has been observed elsewhere \cite{tsiprasRobustnessMayBe2019}. 

Much of the previous adversarial attack literature uses small perturbations that change how a data point is classified \cite{szegedyIntriguingPropertiesNeural2014}. In the image domain, these changes are often so small as to be practical imperceptible. However, log data is more discrete than images, and it is generally not possible to make any change that would be imperceptible. Hence, we do not focus on small perturbations, but rather on generating log lines that are classified as not suspicious. Of course, the generative model can be tasked to match certain features of a real data point, such as a particular user name or timestamp, but that is not something we explore here. 

Relative to the previous literature on adversarial attacks on machine learning cyber defenses, we provide three contributions. First, we develop a framework specific to anomaly detection systems that analyze logs. Previous work focuses on other types of largely numerical data, such as \cite{linIDSGANGenerativeAdversarial2019,usamaGenerativeAdversarialNetworks2019} that look at network flow data with most features being binary zero or one. Our use case requires us to leverage natural language processing approaches to deal with log data. Our second contribution is our finding that robustness against adversarial attack increases with the level of dropout regularization, with a large jump at 90\%. Our third contribution is showing how an adversarial attack can proceed when the attacker does not have access to the data the detector was trained on, nor any close substitute. Much of the prior work allows access to this data to train a surrogate detector model, which is then used to train the generative model.

\section{ADVERSARIAL ATTACK ALGORITHM}
In this section, we describe our adversarial attack algorithm. There is a given black-box detector $D$ that maps inputs to anomaly scores. Our algorithm constructs a generator $G_\theta$ that deceives $D$, in the sense that $G_\theta$ generates fake data that is assigned low anomaly scores by $D$. The success of $G_\theta$ at this deception can be measured using standard metrics on the anomaly score's ability to distinguish fake from real data, as we will discuss in more detail in Section 3. 

The inputs to $D$ are some form of sequence data. Our motivating example is log lines, but our framework could equally be applied to other types of  sequential data such as network packet captures. Inputs are of the form  $X=(x_1,...,x_T)$, where $x_t$ is the $t^{th}$ ``word" in the sequence. The length $T$ is allowed to vary between examples. Each sequence starts with a ``start of line" token. The sequence may also end with an ``end of line" token, which we interpret to mean the sequence is complete. We will also deal with partial sequences that do not have an end of line token. These tokens do not need to be in the raw input data, but can be added as needed. For sequence $X$, we will use $X_{1:t}$ to mean the sub-sequence containing the first $t$ words of $X$.

$G_\theta$ is any neural network sequence generation model with parameters $\theta$. For any sequence $X=(x_1,...,x_t)$, the generator assigns a distribution over the next word in the sequence. We use $G_\theta(x_{t+1}|X)$ to indicate the probability that the next word is $x_{t+1}$ given sequence $X$. The distribution includes the possibility of the ``end of line" token indicating the sequence is complete. By recursively drawing words from this distribution until the end of line token is reached, $G_\theta$ can be used to finish any partial sequence. $G_\theta$ can also be used to make a brand new sequence by starting from the start of line token. 

Our framework does not specify the details of $G_\theta$, such as the architecture or whether it operates on a character or word level. This allows for any sequence generation neural network model to be used. We do require that $G_\theta$ is compatible with backward propagation. Section \ref{exp} provides details on the specific model we use in our experimental setup. 

$G_\theta$ is trained using deep reinforcement learning on the feedback given by the detector's anomaly scores. Our approach uses policy gradient methods and the REINFORCE algorithm \cite{suttonPolicyGradientMethods2000,williamsSimpleStatisticalGradientfollowing1992}, similar to how these techniques have been applied to sequence data in the natural language processing domain \cite{liAdversarialLearningNeural2017,yuSeqGANSequenceGenerative2017}. The mathematical foundations for our approach can be found in those cited works and is therefore omitted here. 

The training procedure lasts for $N$ steps. Each step proceeds as follows. First, $G_\theta$ is used to generate a new complete sequence $X=(x_1,...,x_T)$. Next, for each $t=1,...,T$, we use $G_\theta$ to generate $r$ complete sequences that start from $X_{i:t}$. The detector is then used to assign an anomaly score to each of these $r$ sequences. We take the average of these scores, which we denote $Q_t$. Note that for $t=T$, the sequence $X_{1:T}$ is already complete, so $Q_T$ will just be the anomaly score assigned to the initial sequence $X$. 

At each step $n$, we calculate an estimate for the average anomaly score given to a new sequence generated by $G_\theta$, which we denote $\bar{Q}_n$.  We do so by taking a rolling average of the anomaly scores assigned to the new sequences from the previous 500 steps. Note that $\bar{Q}_n$ is calculated using the brand new sequences generated at the start of each step and not the $r$ sequence completions. At the start of training, $\bar{Q}_1$ is initialized by generating five hundred sequences using the initial generator. 

The difference $Q_t-\bar{Q}_n$ provides an estimate of how suspicious the detector finds the partial sequence $X_{1:t}$, as compared to an average sequence generated by $G_\theta$. We use $Q_t-\bar{Q}_n$ as an estimate of $G_\theta$'s loss for adding $x_t$ to the sequence $X_{1:t-1}$. Therefore, we update the parameters $\theta$ using the following formula: 
\begin{equation}
\theta \leftarrow \theta - \alpha \sum^{T}_{t=1}(Q_t - \bar{Q}_n)\nabla_\theta \log(G_\theta(x_t|X_{1:t-1})+\varepsilon)
\end{equation}
 The parameter $\alpha>0$ is the learning rate, and the parameter $\varepsilon$ is a small value used to avoid taking the log of zero. We used $\varepsilon= 10^{-7}$, but any sufficiently small strictly positive value would work.
 
 Including the $\bar{Q}_n$ term was crucial to making the adversarial attack work. This type of term was used in some previous works using reinforcement learning on natural language (such as \cite{liAdversarialLearningNeural2017}), but not others (such as \cite{yuSeqGANSequenceGenerative2017}). Without the $\bar{Q}_n$ term, the generator either did not improve or saw the loss oscillate during training. This is likely due to the fact that the magnitude of $Q_t$ varies quite a bit over the course of training. Without $\bar{Q}_n$, this would lead to very different loss magnitudes that are not accounted for by one learning rate. Either the training will be too fast when $Q_t$ is large leading to oscillation, or training will be too slow when $Q_t$ is too small leading to no improvement. Of course, this could be corrected by a dynamically adjusted learning rate, but the addition of $\bar{Q}_n$ is simpler and more directly addresses the problem. 
 
 The updating rule described in Equation 1 can be straight-forwardly modified to use a more sophisticated optimization procedure, such as the Adam optimizer \cite{kingmaAdamMethodStochastic2017}. Also, it should be noted that the gradient is not applied to $Q_t$. In other words, $Q_t$ is treated as a constant even though the parameters $\theta$ do influence its value since $Q_t$ is based on sequence completions done by $G_\theta$. This is in line with the REINFORCE algorithm and the mathematical justification for this can be found in prior work  \cite{suttonPolicyGradientMethods2000,yuSeqGANSequenceGenerative2017}.

$G_\theta$ is pretrained on a set of sequence data $\mathcal{X}_{pre}$. As noted in the introduction, $\mathcal{X}_{pre}$ is assumed to only loosely resemble the data that the detector is trained on, and is not sufficient on its own to train a useful generator. The pretraining is only needed so that $G_\theta$ can learn the basic structure of the data, such as an estimate of sequence length and a vocabulary that least partially overlaps with the vocabulary the detector expects. The pretraining is done using the standard next token prediction task using cross entropy loss and the Adam optimizer. 

\begin{algorithm}
	\caption{Adversarial Attack Algorithm}

	\begin{algorithmic}[1]
		\REQUIRE Detector $D$;  Generator model $G_\theta$; number of steps $N$; number of sequence completions $r$; learning rate $\alpha$; pretrain dataset $\mathcal{X}_{pre}$.
		\STATE Initialize $G_\theta$ with random weights $\theta$
		\STATE Pretrain $G_\theta$ on $\mathcal{X}_{pre}$
		\STATE Initialize $\bar{Q}_1$ by generating 500 sequences from $G_\theta$. 
		\FOR{n in 1:N}
			\STATE Generate a sequence $X = (x_1,...,x_T)\sim G_{\theta}$

				\FOR{t in 1:T}
				\STATE Using $G_\theta$, generate $r$ distinct sequence completions starting from $X_{1:t}$  
				\STATE Use $D$ to assign an anomaly score to each completed sequence and calculate an average score $Q_t$.
			
				\ENDFOR
			\STATE Update $\theta$ using Eq. 1.
			\STATE Calculate $\bar{Q}_{n+1}$

		\ENDFOR
	\end{algorithmic}
\end{algorithm}

\section{EXPERIMENTS}\label{exp}
We used publicly available data to validate our adversarial attack algorithm and to analyze whether robustness to that attack varies with the level of dropout regularization. Section \ref{exp_data} describes the dataset used. Sections \ref{exp_detector} and \ref{exp_generator} respectively describe the detector and generator models. Those sections also describe how those models were trained. Section \ref{exp_dropout} examines the impact of dropout on adversarial robustness. 

\subsection{Data}\label{exp_data}
\begin{figure}	
	\centerline{\includegraphics[width=\linewidth]{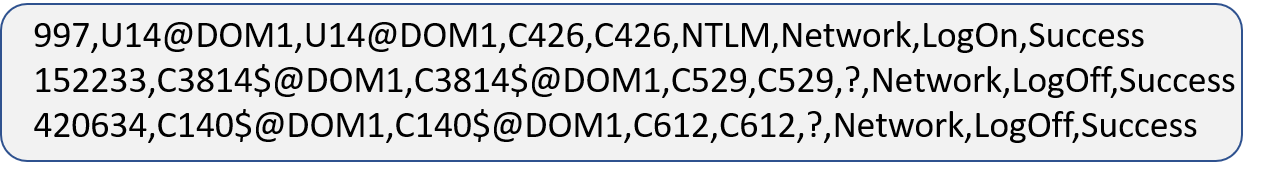}}
	\caption{Example Log Lines from the LANL Data}
	\label{example-lines}
\end{figure}

Experiments were performed using publicly available data from Los Alamos National Lab (LANL) \cite{kentComprehensiveMultiSourceCyberSecurity2015,kentCybersecurityDataSources2015}.\footnote{Data available at https://csr.lanl.gov/data/cyber1/.} The data includes 58 days of authentication logs from their internal computer network. Each day has between 4 and 8 million log lines. A few example log lines from this dataset are displayed in Figure \ref{example-lines}. 

Each log line has comma separated fields in the following order: ``time, source user, destination user, source computer, destination computer, authentication type, logon type, authentication orientation, success/failure." The data was anonymized by replacing values with unique tokens, e.g., U14 replaces a user name. A small number of individual log lines ($<$.001\%) are marked as involved in red team activity. We did not use these labels for training, but only to provide an initial evaluation of our detector models before putting them in the adversarial attack algorithm. 

Of the 58 days, we used data from days 8, 9, and 50. Day 8 is used to train the detector models. Day 9 was used as a test set to provide an initial measure of how well the detector models were performing. The choice of separate days for train and test follows work by Brown et al. \cite{brownRecurrentNeuralNetwork2018} and was done to emulate a cyber security workflow where data from a previous day is used to evaluate current activity. Data from day 9 was also used when evaluating the detector's ability to distinguish real data from fake data created by the generator. 

Lastly, half a percent of the day 50 data was used as the pretrain data for the generator. We added noise by randomly shuffling the data within each column. For example, an entry in the "source computer" field would still be in that field, but assigned to a random row. This shuffling was done to match the assumption that the pretrain data gives some basic structure and vocabulary of the logs, but is too noisy to train a useful generator on its own. The relatively small size and the fact that we use a data from a distant day is also helpful in this regard.

\subsection{Detector Model}\label{exp_detector}
For the detector models, we followed the training method in Brown et al. \cite{brownRecurrentNeuralNetwork2018} who also worked with the LANL dataset.\footnote{Brown et al. present two categories of models: the ``event model" and the ``tiered event model". We used the former.} However, we did not use an LSTM model as they did, but instead used the encoder portion of the transformer model  \cite{vaswaniAttentionAllYou2017}. We did this to sharpen the contrast with our generator model architecture. We trained six different detector models that differed only in their level of dropout probability. The dropout probability levels used were 0\%, 10\%, 30\%, 50\%, 70\%, and 90\%.

The transformer encoder we used had 8 heads, 1 layer, and a model dimension of 512. Additional layers were experimented with, but did not meaningfully impact the detector's performance, so one layer was used to speed training time. The output of the transformer encoder was passed to a fully connected layer with one node for each word in the detector's vocabulary. The detector's vocabulary was determined by the training data with a word needing to appear 40 times in the dataset to be included. The vocabulary also included an ``out of vocabulary" token for unknown words.  

We trained the detector to predict the next word in each log line using the words that came before. To do so the words from $i+1$ onward were masked when predicting the $i^{th}$ word. Cross entropy loss and the Adam optimizer  $(\beta_1,\beta_2=.9,.999)$  were used with a learning rate of $10^{-4}$. The summed cross entropy loss for each word of a log line was used as the anomaly score. This captures how unusual the detector finds the line. Each model was trained over four epochs on day 8 of the LANL dataset, which was enough to converge without over fitting. 

As an initial benchmark of performance, we used day 9 data to calculate test loss and ability to detect red team activity. The ability to detect red team activity was measured by the AUC (Area Under the ROC Curve) when using the anomaly scores to predict the red team labels. Table 1 reports these results. Notice that performance generally declines as dropout probability increases, with a notable drop in performance at 90\% dropout. Hence, from this evaluation the 90\% dropout model appears suboptimal. 

\begin{table}
	
	\begin{center}
		{\caption{Detector Performance by Dropout}\label{detector_benchmark}}
		\begin{tabular}{ccc}
			\toprule
			Dropout Probability & Red Team Detection AUC &      Loss \\
			\midrule
			0\% &  0.991 &  1.898 \\
			10\% &  0.992 &  1.934  \\
			30\% &  0.992 &  1.994 \\
			50\% &  0.990 &  2.327 \\
			70\% &  0.987 &  2.378 \\
			90\% &  0.933 &  3.206 \\
			\bottomrule
		\end{tabular}
	\end{center}
\end{table}

\subsection{Generator Model}{\label{exp_generator}

Our generator model used a Long Short-Term Memory (LSTM) architecture \cite{hochreiterLongShortTermMemory1997}. The log lines were fed into a word embedding layer of dimension 256 which were then passed to an LSTM with hidden size 256. At each step of the LSTM, the output was passed through a dropout layer with drop probability of 50\%, and then a fully connected layer that predicts the next word. A softmax operator is used on the values of the fully connected layer to generate a distribution over the next word. 

The model was pretrained on one half of one percent of day 50 data with shuffled columns, as described above. The data was split into 80\% train set and 20\% test set. The pretain dataset was used to set the vocabulary of the generator model, which determines the number of embeddings used on the input and the number of possible output words. The pretraining lasted for 20 epochs, which was determined by an early stopping rule based on when the test loss started to increase. Cross entropy loss was used with a learning rate of $10^{-4}$ and the Adam optimizer $(\beta_1,\beta_2=.9,.999)$. 

Next, the generator was trained using the adversarial attack algorithm described in Section 2. We repeated this process six times, once for each of the six detectors with the varying levels of dropout. We used a learning rate of $\alpha=10^{-5}$, the number of sequence completions per step was set as $r=50$, and the number of steps was $N=100,000$. 

Every 2,500 steps, 20,000 log lines were generated and a mean anomaly score was calculated.  These means are shown in Figure \ref{adverse_training} for the six detector models. The dotted  line in each figure gives the average anomaly score assigned by that detector model on real log lines from day 9 of the LANL dataset. Figure \ref{adverse_training} demonstrates that the adversarial training successfully improves the generator's ability to fool the detector model in the sense of decreasing the mean anomaly score of the fake log lines. In every case, except for 90\% dropout, these means are near or below the mean anomaly score of real data.  Also, note that in each case the average anomaly score for the fake data starts much higher than the dotted lines. This indicates that the generator models are not effective with just the pretraining, validating our assumption that the pretrain data is not useful on its own.

\begin{figure}
	\centerline{\includegraphics[width=\linewidth]{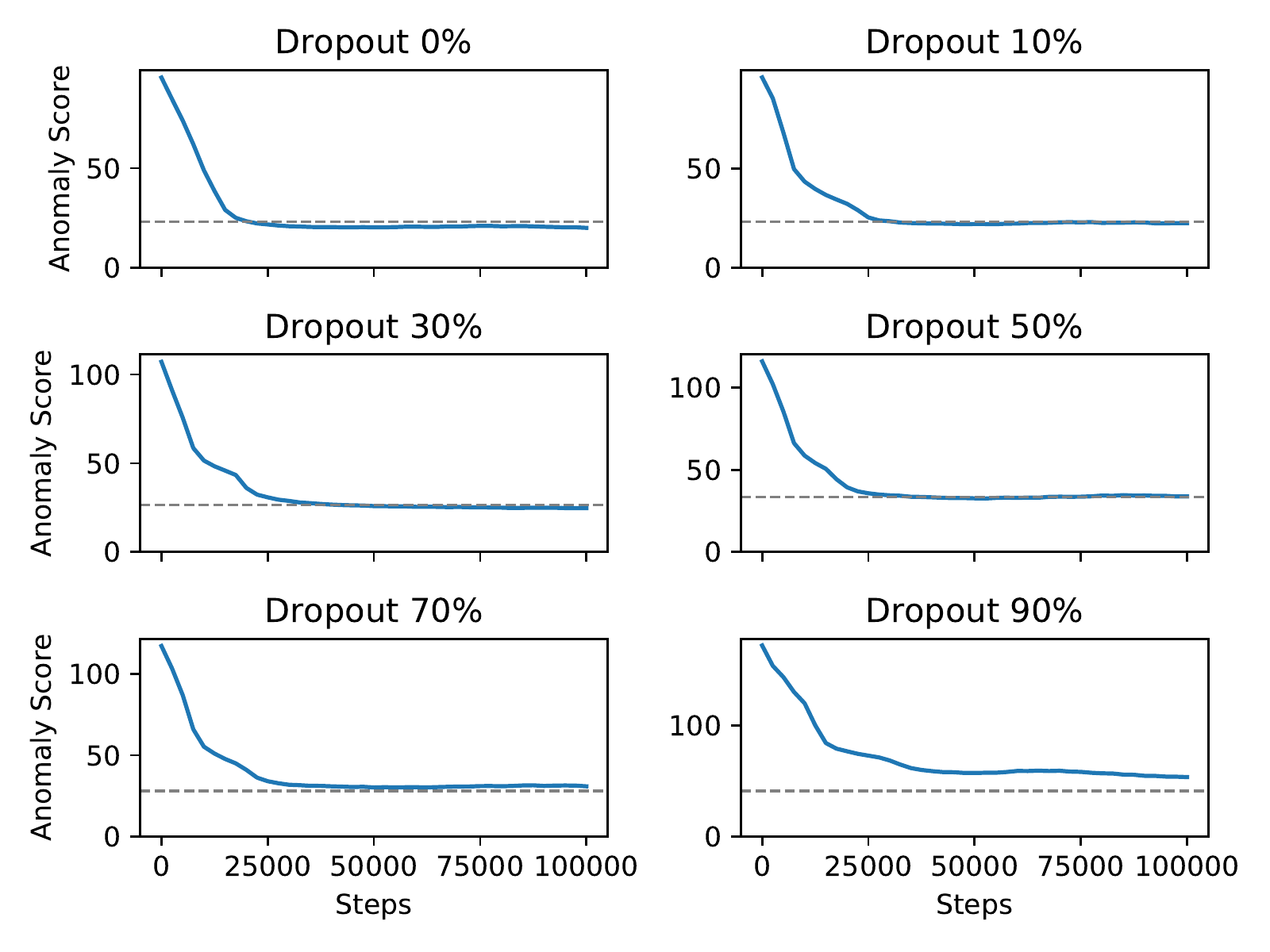}}
	\caption{Average anomaly scores during the adversarial training process}
	\label{adverse_training}
\end{figure}

Notably, these average anomaly scores are not comparable across the different models. Different detector models assign different distributions of anomaly scores, as can be seen by the different levels of the dotted in each graph of Figure \ref{adverse_training}. In the next section we use a different metric to compare across dropout levels, which we turn to now.

\subsection{Dropout Impact}\label{exp_dropout}
We now look at how the dropout level of the detector model impacts robustness to adversarial attack. To measure this robustness, we calculate the AUC score when using the detector's anomaly score to distinguish which log lines are real and which are fake. The AUC score provides a comparable metric across the different detector models, since it depends only on the ordinal rank of the anomaly scores. Moreover, the AUC score is sensitive to the entire distribution of anomaly scores instead of just capturing one element such as the mean. 

Table \ref{dropout_impact_auc} shows the AUC scores for the varying levels of detector dropout. The AUC score was measured at the end of the 100,000 steps of adversarial training. A dataset of 40,0000 examples was used, consisting of half data drawn from day 9 of LANL dataset and half fake data created by the generator model. While not displayed in Table \ref{dropout_impact_auc}, the initial AUC score for each model was very close to 1 ($> .99$) for each of the six detector models, which further validates the assumption that the pretraining cannot deliver a useful model. 

Table \ref{dropout_impact_auc} shows a distinct trend of higher dropout probability leading to higher robustness against adversarial attack, as measured by a higher AUC score. The only exception is the 30\% dropout probability having a higher AUC score than 50\%. The highest AUC score by a significant margin occurs at 90\% dropout, which is also the only AUC score greater than .5. An AUC of .5 is what could be achieved by a model guessing randomly. Hence, the 90\% dropout is the only dropout level that is truly effective at warding off the adversarial attack. 

A dropout level of $90\%$ is much higher than typically seen in transformer models. For example the paper that introduced the transformer model \cite{vaswaniAttentionAllYou2017} used dropout between $10\%$ and $30\%$. It is notable that dropout $90\%$ had the worst baseline performance when there was no adversary, as we saw in Table 1. This suggests a trade-off between baseline performance and adversarial robustness, which is in line with what has been observed elsewhere \cite{tsiprasRobustnessMayBe2019}.

\begin{table}
	
	\begin{center}
		{\caption{AUC Scores by Dropout}\label{dropout_impact_auc}}
		\begin{tabular}{ccc}
			
			\toprule
			Dropout Probability & AUC Score  \\
			\midrule
			0\%  & 0.165    \\
			10\% &  0.272  \\
			30\% &  0.329   \\
			50\% &  0.277 \\
			70\% &  0.422  \\
			90\% &   0.680  \\
			\bottomrule
		\end{tabular}
	\end{center}
\end{table}

A common pitfall with this type of adversarial training is that the generator might collapse into producing the same log line over and over again. The AUC score would not necessarily detect this, since that one log line could have a low anomaly score and hence pass detection. Figure \ref{dropout_impact_unique_counts} shows the percentage of duplicates present in the 20,000 generated logs lines used to calculate the mean anomaly scores seen in Figure 2. More precisely, the y-axis of the figure is the percentage of logs lines that would need to be removed in order to make the set of logs lines fully unique. 

The amount of duplicate log lines does increases as training goes on, but at no point is the duplicates percentage more than 30\%. Moreover, higher dropout models do not exhibit a higher duplicate percentages. In fact, the 90\% dropout model has a notably lower duplicate percentage for most of the training process. This shows that the higher AUC of the higher dropout models is not due to those models creating more duplicates.

\begin{figure}
	\centerline{\includegraphics[width=\linewidth]{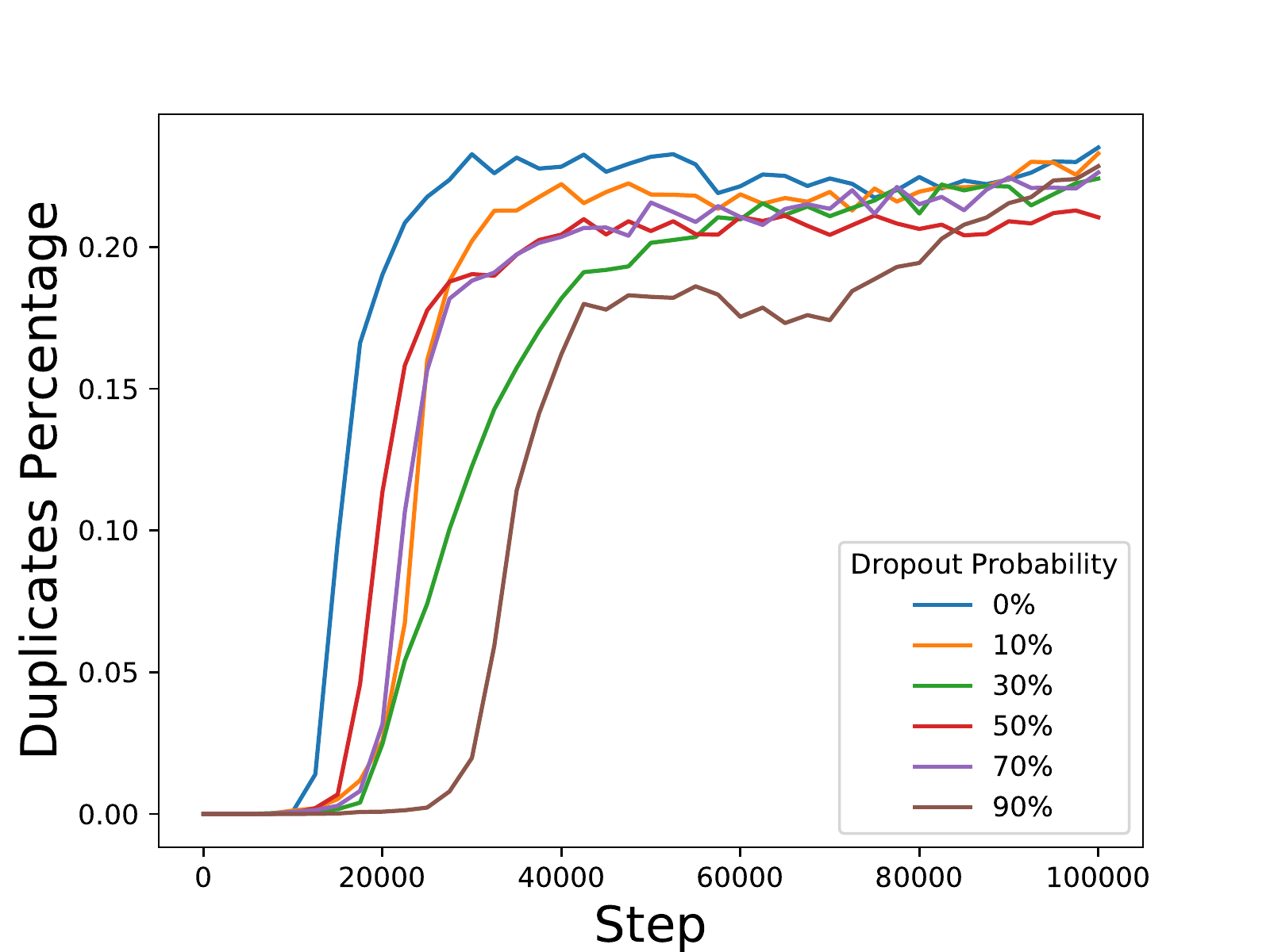}}
	\caption{Percentage of duplicate log lines generated over the course of training}
	\label{dropout_impact_unique_counts}
\end{figure}

We also recalculated the AUC scores after forcing the generator model to create 20,000 unique log lines. This was done by  continuously generating log lines until a set of 20,000 fully unique log lines had been created. The AUC score was then calculated in the same way as before using the 20,000 unique log lines. The result are shown in Table \ref{dropout_impact_auc_unique}. These results are very similar to Table \ref{dropout_impact_auc}, which again shows that model collapse was not an important feature of our results.

\begin{table}
	
	\begin{center}
		{\caption{AUC Scores by Dropout with Unique Dataset}\label{dropout_impact_auc_unique}}
		\begin{tabular}{ccc}
			\toprule
			Dropout Probability & AUC Score \\
			\midrule
			0\% &  0.169   \\
			10\% &  0.273    \\
			30\% &  0.333   \\
			50\% &  0.274   \\
			70\% &  0.428   \\
			90\% &   0.679 \\
			\bottomrule
		\end{tabular}
	\end{center}
\end{table}

\section{DISCUSSION}

In this paper, we developed a way to test the adversarial robustness of machine-learning based cyber defenses. Our framework studies the case where an adversary has captured a black-box copy of an anomaly detection model on log data. We use a publicly available dataset to show that this type of defensive approach is in fact vulnerable to attack. We also apply our framework to evaluate the impact of dropout levels on the detector model's robustness against an adversarial attack. We find that robustness increases with the drop probability, with a significant jump occurring at 90\%, which is well above the typical range. 

Dropout regularization is often understood as effectively creating an ensemble of different models within a single neural network \cite{hintonImprovingNeuralNetworks2012}. This provides a possible explanation for why dropout is effective in increasing adversarial robustness. It is naturally harder to reverse engineer the decision rule of an ensemble of models as opposed to a single model, which makes the generator's task more difficult. This suggests the possibility of using an actual ensemble of models to increase adversarial robustness. Another natural avenue to explore is other types of regularization besides dropout, such as L2 or early stopping. We view our results here as only a starting point for this investigation, with many more possible avenues open for future research. 

Another potential direction for future work is whether our adversarial attack framework could be modified to dispense with the pretrain data entirely. This would answer whether an attack could succeed against a system where the attacker has no frame of reference for what the data looks like. We hypothesize that doing such an attack would require an initial step to discover basic structural facts about the data, such as a rough range of sequence lengths and a partial vocabulary. The feasibility of such a step depends on whether the feedback from the detector makes fine enough distinctions. For example, does the detector differentiate between an input where only one of the token is in the detector's expected vocabulary versus an input that has no tokens in the detector's expected vocabulary. A detector that makes fine enough distinctions would score both inputs as anomalous, but the latter as being more so. The detector model we used in our experiments does make this type of distinction, but others may not.

In this paper, we focused on the case of attacking anomaly detection models that use log data. However, our framework is more general than that. The detector can be any model that returns feedback on how suspicious it finds each input. The detector does not even need to be a machine learning model. For example, a rules-based system that returns either alert or no alert on each input could be used in our framework. Additionally, the input data does not have to be log data, but can be any type of sequence data, which includes network packet captures or command-line inputs by a user. Log data is a a good fit with the natural language techniques we used, but those techniques are not limited to that use case.   

We conclude by noting that our approach draws inspiration from the software testing domain, where code is rigorously tested for problematic inputs that cause bugs or crashes. Of course, those problematic inputs are precisely what malicious attackers are looking to find, so deliberately uncovering them poses a certain amount of risk. However, it is generally considered better to know about such potential vulnerabilities so that they can be addressed rather than simply hope that attackers will not discover them. This is the same perspective that we take in our work here. Our adversarial attack algorithm uncovers an effective method defensive detector. However, we do so in service of searching for how to best prepare against such an attack. 

\bibliography{ecai,ML}
\end{document}